\documentclass[letterpaper, 10 pt, conference]{ieeeconf}  

\IEEEoverridecommandlockouts                              

\usepackage{amsmath} 
\usepackage{amssymb}  

\usepackage{cite}
\usepackage{graphicx}
\usepackage{booktabs}
\usepackage{threeparttable}
\usepackage[table]{xcolor}
\usepackage{makecell}
\usepackage[caption=false,font=footnotesize]{subfig}
\usepackage{algorithm}
\usepackage{algpseudocode}

\usepackage{xspace}
\newcommand{\etal}{\textit{et~al.}\@\xspace}

\title{\LARGE \bf
Commonsense-Grounded Path Planning from Abstract Instructions
}

\author{Masafumi Endo$^{1}$, Kohei Honda$^{1, 2}$, Ryo Yonetani$^{1}$
\thanks{This work was supported by JSPS KAKENHI Grant Number 26K02991.}%
\thanks{This work has been submitted to the IEEE for possible publication. Copyright may be transferred without notice, after which this version may no longer be accessible.}%
\thanks{$^{1}$CyberAgent Inc., Tokyo, Japan\quad $^{2}$Nagoya University, Aichi, Japan}%
\thanks{{\tt\small \{endo\_masafumi, honda\_kohei, yonetani\_ryo\}@cyberagent.co.jp}}%
}

\begin{document}

\bstctlcite{IEEEexample:BSTcontrol}

\maketitle
\thispagestyle{empty}
\pagestyle{empty}

\begin{abstract}

We present \emph{commonsense ranked search} (CoRS), a novel path planner that turns an abstract instruction into a route that follows commonsense.
While existing methods respect the considerations written down in advance, a robot working among people must follow those left unstated too, as with a wet floor that a worker avoids without being told.
CoRS leverages large language models (LLMs) and vision-language models (VLMs) as commonsense knowledge to reason about these latent considerations in its planning. 
Given an abstract instruction (\emph{e.g.}, ``move carefully'') and visual observations of each region in the environment, CoRS derives a consideration for each region, as in ``this wet floor is slippery and worth a detour.''
It then compares the considerations between regions to see which of the two the robot should avoid more, as in ``the crowd is worse than the wet floor.''
These judgments sort the regions into a commonsense ranking, whose costs drive a conventional search that always returns a valid route.
We build a benchmark for planning under latent considerations, with three environments, 1350 problems, and five instructions at three levels of abstraction.
Experiments show that CoRS discovers the unstated considerations and goes around the ones worth a detour while crossing the rest, a behavior that recent LLM-based planners do not achieve.

\end{abstract}


\section{Introduction}
\label{sec:introduction}

Society expects us to act with commonsense. 
Imagine a supermarket worker is asked to ``carry this basket to the backroom carefully.'' 
Tactile paving lines the floor, customers queue at the checkout, and an aisle is wet after cleaning. 
The worker hears nothing more than \emph{carefully}, yet adjusts the route so that each of them gets the right amount of care.
Commonsense carries us from abstract instructions to concrete decisions, on which everyday life runs smoothly.

\begin{figure}[t]
    \centering
    \includegraphics[width=\linewidth]{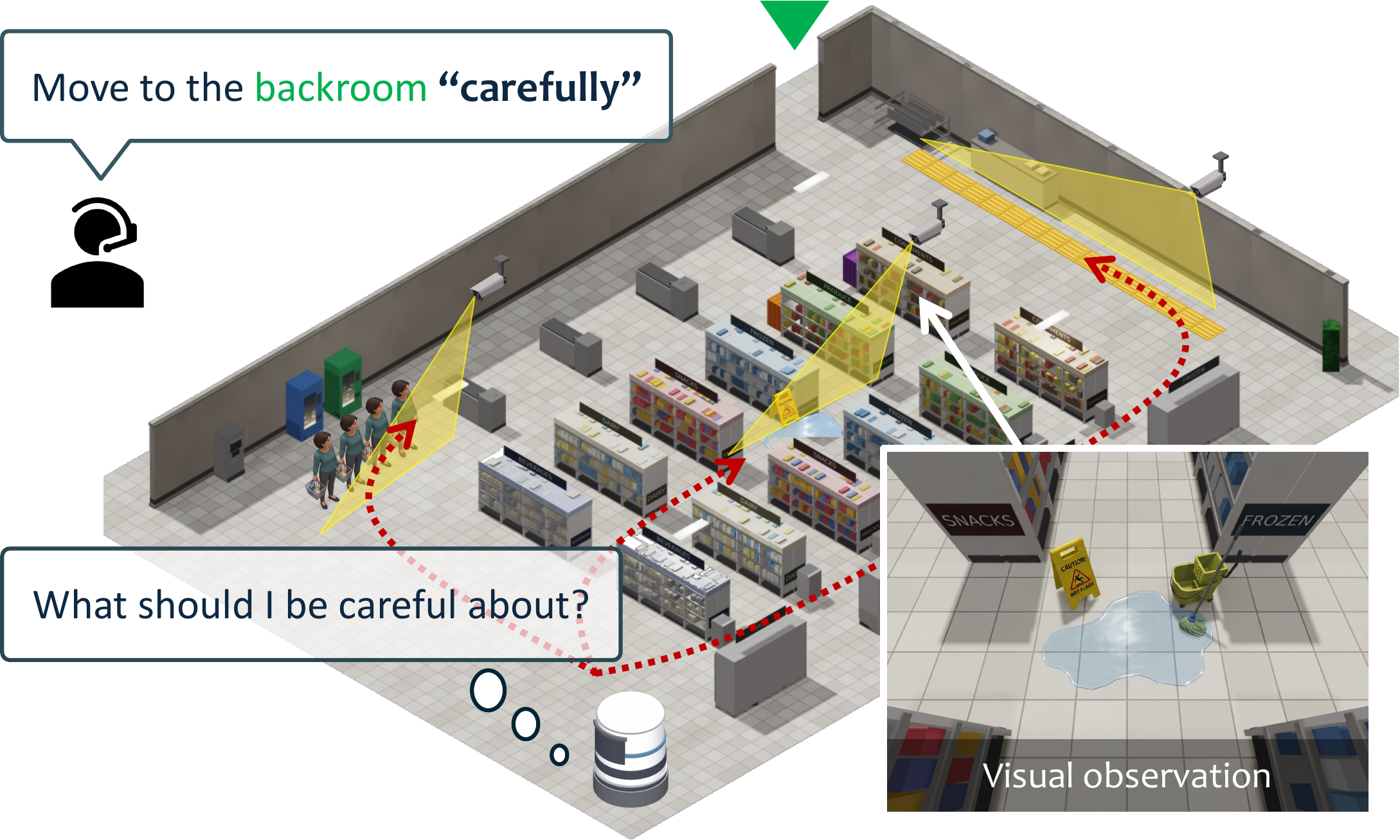}
    \caption{\textbf{Motivating scenario: Planning by commonsense.} 
    In our simulator, a robot is asked to move \emph{carefully} in a supermarket and sees each region through a surveillance camera. The instruction does not say which regions to avoid. No route avoids all of them, so it must decide which to cross.}
    \label{fig:teaser}
\end{figure}

Our goal is to make mobile robots move among people with such commonsense.
These robots have to turn the given instruction into a route (Fig.~\ref{fig:teaser}).
Each situation on its way puts a \emph{consideration} on the robot: risk, constraint, or preference.
Existing work has incorporated such considerations into planning as costs translated from instructions~\cite{weerakoon2025behav,huo2026norm,bao2025path}, regions on the map that language can refer to~\cite{shah2022lmnav,huang2023visual}, or rules with priorities set in advance~\cite{tumova2013minimum,tumova2013least,castro2013incremental,censi2019liability}.
However, these methods assume every consideration to be written down explicitly, rather than latent behind an abstract instruction.

We thus propose \emph{commonsense-grounded path planning}, a novel planning problem for a robot that follows an abstract instruction.
Given the instruction and a collection of visual observations that cover multiple regions across the environment, the robot must find the right route without explicit considerations to respect.
The challenge is that the considerations change with the instruction, the task, and the remaining routes together.
Suppose the robot finds a wet aisle on its route.
Usually it can cross, but with a cup of coffee to deliver it should take a detour.
When no detour exists, it should cross rather than abandon the task.
Large language models (LLMs) and vision-language models (VLMs) can judge such situations at planning time~\cite{weerakoon2025behav,fang2026obstacles}. 
Yet the model may write a route that the map does not have~\cite{chen2024mapgpt,zhou2024navgpt,zhou2024navgpt2}, or select among the candidates a classical planner prepared from geometry alone~\cite{song2025guide,fang2026obstacles}.

As a solution to this problem, we propose \textbf{commonsense ranked search (CoRS)}.
Given an abstract instruction and visual observations, it leverages LLMs and VLMs to infer the latent considerations and turn them into the costs of search-based planning. 
It first describes the observation of each region as text through a VLM.
It then puts two questions to an LLM: whether the region is worth a detour and which of two regions the robot should avoid more.
The answers sort the regions against one another and against the length of the route into a \emph{commonsense ranking}.
Its costs drive a classical search and keep every route valid.
The same region is thus avoided in one situation and crossed in another, as the instruction, the task, and the remaining routes change.

We evaluate CoRS in simulated everyday environments where service robots work, such as supermarkets and hospitals.
We build a \emph{commonsense planning benchmark}, a simulator and a problem set for evaluating LLM-based planners under latent considerations.
We design the situations ourselves and record the right behavior for each, which existing environments do not provide.
Our benchmark covers three environments and 1350 problems, where five instructions at three levels of abstraction vary how much the robot works out for itself.
Experiments show that CoRS finds the unstated considerations and goes around the ones worth a detour with far fewer needless detours than LLM-based planners.

Our contributions are as follows:
\begin{itemize}
    \item \textbf{Commonsense-grounded path planning.} A new planning problem where no consideration is stated, unlike the settings that give the rules to follow in advance.
    \item \textbf{Commonsense ranked search.} A planner that brings an LLM into the cost of the search to weigh the regions relative to one another rather than in isolation.
    \item \textbf{Commonsense planning benchmark.} A simulator and a problem set that record the right behavior under latent considerations, which no existing environment provides. Fig.~\ref{fig:teaser} shows one of its environments.
\end{itemize}
We will release the benchmark with its simulator and baselines upon publication.
\section{Related Work}
\label{sec:related_work}

We review commonsense in robot navigation, instruction-guided costs, and foundation models as planners.

\paragraph{Commonsense in robot navigation}
Robot navigation among people involves not only geometric obstacles but also the considerations that hold in a place.
Classical work enumerates them by hand, along with the priorities that decide which to break first~\cite{tumova2013minimum,tumova2013least,castro2013incremental,censi2019liability}.
Sathyamoorthy \etal classify the scene with a VLM, but still write the behavior for each scene by hand~\cite{sathyamoorthy2024convoi}.
RoboGuard grounds the rules written down on the scene with an LLM~\cite{ravichandran2026safety}.
Yet each rule is checked as a binary condition, so the robot cannot weigh how much it breaks one rule against another.
Recent work instead learns them from how people already move.
Choi \etal train a navigation policy on hours of human driving~\cite{choi2025canvas}.
Chen \etal learn from expert trajectories as well, but add a reward that scores social compliance~\cite{chen2026socialnav}.
However, a robot trained this way cannot find the unstated considerations, and a recent benchmark reports far lower compliance with them than with the stated goals~\cite{zhao2026normact}.

\paragraph{Instruction-guided cost}
When the instruction states the rules to follow, a planner can encode them as the cost on the map and search for a suitable route.
One approach turns the instruction into a cost, but the map holds geometry alone~\cite{bao2025path}.
Other work derives the required behavior from the instruction, asks a VLM to check each place against it, and raises the cost where the place fails~\cite{weerakoon2025behav,huo2026norm}.
A score from the model can also be the cost, as LFG scores each frontier for object-goal navigation~\cite{shah2023navigation} and Song \etal score the social entities around the robot for a local planner~\cite{song2024vlm}.
Sugino \etal bring it to the instruction, where a VLM scores each observation image on a graph and the score scales the edge costs~\cite{sugino2026vang}.
While these methods cover a range of instructions, they judge each place on its own, so a place carries the same cost in any situation.

\paragraph{Foundation models as planners}
Foundation models carry commonsense about the behavior that suits a place, and recent work asks them to plan the route with it.
MapGPT and the NavGPT family write the places and their connections into the prompt, from which the model moves among the neighbors at each step~\cite{chen2024mapgpt,zhou2024navgpt,zhou2024navgpt2}.
Yet even in a small maze, an LLM writes an invalid route due to misreading the spatial relations, such as one through an obstacle or into a dead end~\cite{zhang2025mitigating}.
Another approach therefore restricts the model to selecting one of the candidate routes from a classical planner~\cite{song2025guide,fang2026obstacles}.
The candidates come from geometry alone, on the explicit assumption that the right route for the situation is among them~\cite{fang2026obstacles}.

In summary, existing work either converts the considerations written down into a fixed cost per place, or leaves the route to a foundation model outside the search.
We instead integrate the model into the search, where it infers the latent considerations and ranks the regions against one another.
\section{Preliminaries}
\label{sec:preliminaries}

We describe standard forms of path planning on a topological map and the costs we build on, and then state the problem we address.

\subsection{Path Planning on Topological Maps}
\label{sec:path_planning}

Let $G=(\mathcal{V},\mathcal{E})$ be an undirected topological map of the environment, where the vertices $\mathcal{V}$ are places and each edge $(u,v)\in\mathcal{E}$ connects places that the robot can move between, with a length $d(u,v)>0$.
A set $\mathcal{R}$ of regions represents the situations on the floor, such as a wet aisle or a queue.
Each region $r\in\mathcal{R}$ covers a part of the map and has one or more observation images $o_r$ that show the situation from above.
A route $P=(v_0,\dots,v_T)$ is a sequence of places from $v_0=v_\mathrm{s}$ to $v_T=v_\mathrm{g}$ with $(v_{t-1},v_t)\in\mathcal{E}$ for all $t$, where $v_\mathrm{s}$ and $v_\mathrm{g}$ are a start and a goal.
With an arbitrary nonnegative cost $w(u,v)$ on every edge, the cost of a route is the sum of the edge lengths and the costs of the edges along it,
\begin{equation}
    J(P)=\sum_{t=1}^{T}\bigl[\,d(v_{t-1},v_t)+w(v_{t-1},v_t)\,\bigr].
    \label{eq:route_cost}
\end{equation}
A classical graph search returns the route that minimizes $J(P)$, so $w$ is the only factor we can control.

\subsection{Consideration Levels as Costs}
\label{sec:priorities_as_costs}

Each region $r$ carries a set $C_r$ of considerations.
A consideration is a demand that the situation there puts on the robot, such as the risk of a wet floor, the constraint of a staff-only area, or the preference for quiet near a resting patient.
When every route to the goal passes through a region with a consideration, the robot has to choose which one to cross.
Each region with a consideration takes a level, given by a function $k : \{r \mid C_r \neq \emptyset\} \to \{1, \dots, K\}$.
Let $e_r(u,v)\geq 0$ be the length of the edge $(u,v)$ inside the region $r$.
The levels then replace the scalar cost in \eqref{eq:route_cost} with a vector $c(P)\in\mathbb{R}^{K+1}$, one component per level and the last for the length,
\begin{equation}
    \begin{aligned}
    c_j(P)&=\sum_{t=1}^{T}\sum_{r:\,k(r)=j} e_r(v_{t-1},v_t)\quad\text{for } j=1,\dots,K,\\
    c_{K+1}(P)&=\sum_{t=1}^{T} d(v_{t-1},v_t).
    \end{aligned}
    \label{eq:level_cost}
\end{equation}
Routes compare on $c(P)$ component by component from the first.
A region of level $j$ thus costs more than any length inside regions of levels $j'>j$ together with any route length.
When every route crosses a region, the search returns the one that crosses the least at the highest level it cannot avoid.
Such costs appear in planning for autonomous driving under traffic rules, where the levels are the priorities among the rules, known as minimum-violation planning and rulebooks~\cite{tumova2013least,castro2013incremental,censi2019liability}.
Those methods count a violation under its rule, but keep both $C_r$ and $k$ fixed in advance.

\subsection{Problem Setup}
\label{sec:problem_setup}

We now formalize \emph{commonsense-grounded path planning}, where a robot finds a route on $G$ under an instruction $\ell$ given in natural language.
A problem instance is a tuple $(G,\mathcal{R},\{o_r\},v_\mathrm{s},v_\mathrm{g},\ell)$, on which we make the following three assumptions.
\begin{enumerate}
    \item The instruction comes at various levels of abstraction a person would use, from an explicit rule such as ``keep off the tactile paving'' to a single word such as ``carefully.'' It may also state the task the robot is on, such as carrying glassware.
    \item The situation in a region is known only through its observation images $o_r$. This holds in indoor environments such as supermarkets, where surveillance cameras already cover the floor.
    \item Neither $C_r$ nor the level $k$ is given. They follow from the situation in each region, together with the instruction, the task, and the routes that remain on the map.
\end{enumerate}
We call \emph{commonsense} the judgment that derives $C_r$ and $k$ from the situation when the instruction leaves them unstated, where a demand on people weighs more than one on the robot.
The solution is a route $P$ on $G$ that follows it.
The standard form cannot produce it, since it takes $C_r$ and $k$ as given.
We instead derive both at planning time, as described next.
\section{Commonsense Ranked Search}
\label{sec:method}

We present \emph{commonsense ranked search} (CoRS), a path planner that derives $C_r$ and $k$ from the instruction and the observations.
CoRS judges whether each region is worth a detour and which of two regions the robot should avoid more.
The answers then form a \emph{commonsense ranking}, a single order over the regions and the length of the route.
A search on the costs from this ranking returns the route.
Algorithm~\ref{alg:cors} summarizes the pipeline.

\subsection{Grounding the Instruction on Each Region}
\label{sec:grounding}

CoRS first reads each region on its own (Algorithm~\ref{alg:cors}, L2--L3).
A VLM writes a text $t_r$ from the observation images $o_r$ of every region.
The text is a neutral description of the observation, such as a queue at the checkout or a cart left in the aisle, for the LLM to judge.
An LLM then reads $\ell$ and $t_r$ and answers whether the region is worth a detour,
\begin{equation}
    a_r = \textsc{WorthDetour}(\ell, t_r) \in \{\texttt{detour},\ \texttt{cross},\ \texttt{none}\},
    \label{eq:worth_detour}
\end{equation}
where \texttt{none} marks a region the text gives no basis to judge.
The LLM writes the reason before each answer, which states the consideration the situation puts on the robot and thus derives $C_r$.
Under ``you carry a full tray'' in Fig.~\ref{fig:problem_example}, the LLM answers \texttt{detour} for the wet floor (``it is slippery'') and \texttt{cross} for the floor sticker (``it is safe to step on'').
This step decides whether a region matters, while the comparison below decides how much.

\subsection{Comparing Two Regions}
\label{sec:comparison}

\begin{figure}[t]
    \centering
    \includegraphics[width=\linewidth]{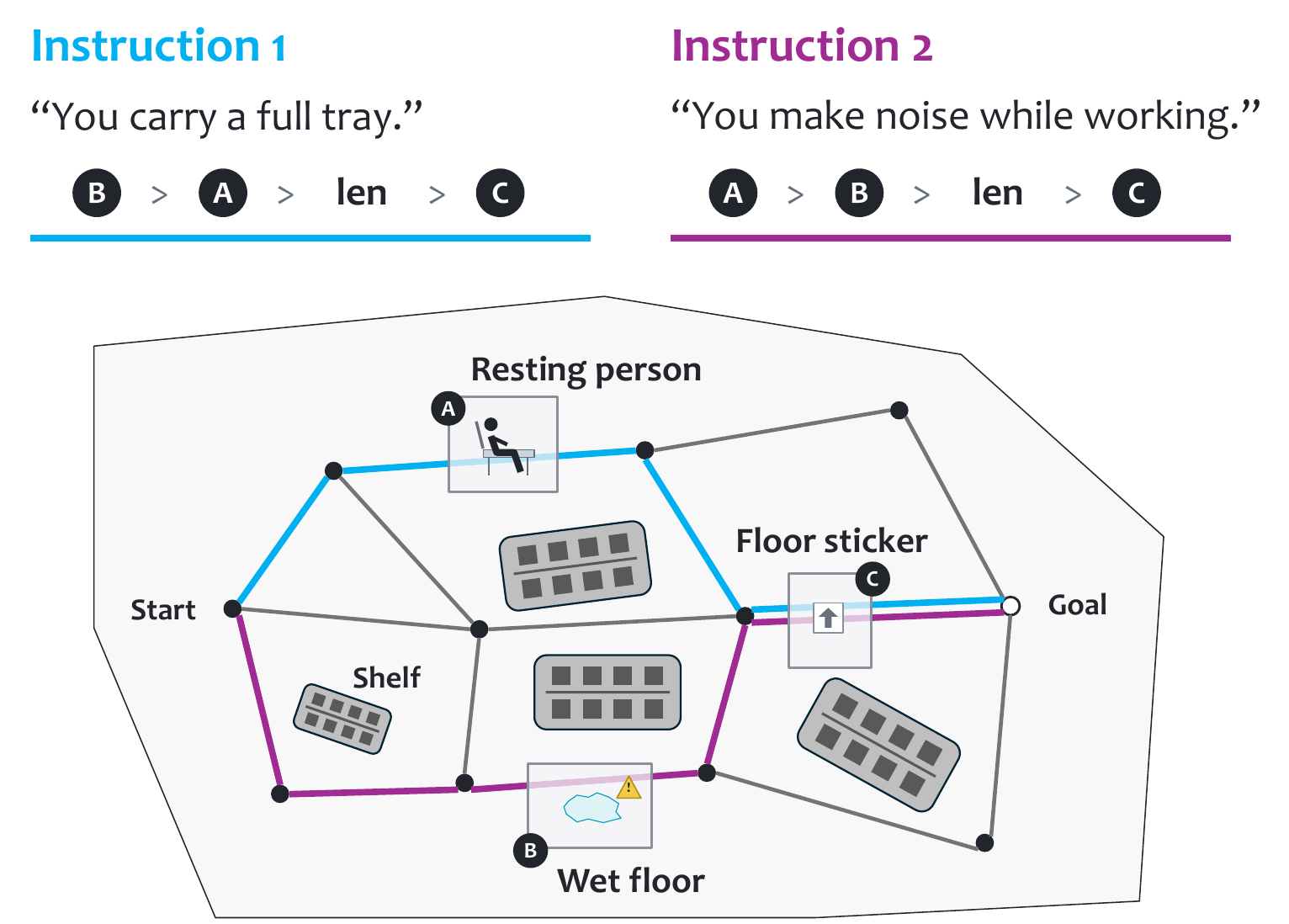}
    \caption{Example problem instance solved by CoRS.
    The same regions get a different ranking and a different route under each instruction.}
    \label{fig:problem_example}
\end{figure}

CoRS also compares the regions with one another.
For a pair $r, r'$, an LLM reads $\ell$ and the two texts and answers which of the two the robot should avoid more,
\begin{equation}
    b_{r,r'} = \textsc{Compare}(\ell, t_r, t_{r'}) \in \{r,\ r',\ \texttt{tie}\},
    \label{eq:compare}
\end{equation}
where \texttt{tie} covers an answer that the two are alike and one that the texts give no basis to separate them.
In Fig.~\ref{fig:problem_example}, the LLM keeps the robot off the resting person under ``you make noise while working'' (``the noise reaches the person'').
It keeps the robot off the wet floor instead under ``you carry a full tray'' (``a spill would be worse than the noise'').
An absolute score instead tends toward caution and leaves little difference among the regions.
An LLM also judges texts more reliably in pairs than by a score alone~\cite{qin2024large,liusie2024llm}.

\begin{algorithm}[t]
\caption{Commonsense Ranked Search}
\label{alg:cors}
\begin{algorithmic}[1]
\Require problem instance $(G,\mathcal{R},\{o_r\},v_\mathrm{s},v_\mathrm{g},\ell)$
\Ensure route $P$ from $v_\mathrm{s}$ to $v_\mathrm{g}$
\For{$r \in \mathcal{R}$}
  \State $t_r \gets \textsc{Describe}(o_r)$ \Comment{VLM}
  \State $a_r \gets \textsc{WorthDetour}(\ell, t_r)$ \Comment{LLM}
\EndFor
\State $\mathcal{L} \gets \textsc{Sort}(\mathcal{R} \cup \{\mathrm{len}\}, \prec)$ \Comment{$\prec$ from \eqref{eq:comparator}}
\State $k \gets$ level of each element in $\mathcal{L}$
\State $P \gets \textsc{Search}(G, v_\mathrm{s}, v_\mathrm{g}, c)$ \Comment{$c$ from $k$ by \eqref{eq:ranking_cost}}
\State \Return $P$
\end{algorithmic}
\end{algorithm}

\subsection{Sorting into a Commonsense Ranking}
\label{sec:ranking}

The two answers above define a comparator over the regions and the length of the route, written $\mathrm{len}$,
\begin{equation}
    x \prec y =
    \begin{cases}
        b_{x,y} & x, y \in \mathcal{R},\\
        a_x & x \in \mathcal{R},\ y = \mathrm{len},
    \end{cases}
    \label{eq:comparator}
\end{equation}
where a region goes above $\mathrm{len}$ when $a_x$ is \texttt{detour} or \texttt{none}, and below it when $a_x$ is \texttt{cross}.
A region with no basis thus stays above $\mathrm{len}$, as the standard form of Section~\ref{sec:priorities_as_costs} places every consideration.
Sorting $\mathcal{R}\cup\{\mathrm{len}\}$ under $\prec$ then returns a commonsense ranking, an order over the regions and the length with ties (L5).
In Fig.~\ref{fig:problem_example}, the floor sticker stays below $\mathrm{len}$ under both instructions, while the resting person and the wet floor swap above it.
The sort queries the LLM only for the pairs it needs, rather than for all $|\mathcal{R}|^2$ of them.

Let the ranking have levels $1,\dots,K$, with $\mathrm{len}$ alone at level $k_{\mathrm{len}}$ and $k$ now defined on every region (L6).
Then \eqref{eq:level_cost} generalizes to a vector $c(P)\in\mathbb{R}^{K}$, one component per level and the length at component $k_{\mathrm{len}}$,
\begin{equation}
    \begin{aligned}
    c_j(P)&=\sum_{t=1}^{T}\sum_{r:\,k(r)=j} e_r(v_{t-1},v_t)\quad\text{for } j\neq k_{\mathrm{len}},\\
    c_{k_{\mathrm{len}}}(P)&=\sum_{t=1}^{T} d(v_{t-1},v_t).
    \end{aligned}
    \label{eq:ranking_cost}
\end{equation}
A region above $\mathrm{len}$ costs more than any detour around it, so the robot goes around it whenever a route exists.
A region below $\mathrm{len}$ matters only between routes of equal length, so the robot passes it whenever the route through it is shorter.

\subsection{Searching on the Ranking}
\label{sec:search}

A search-based planner on $c(P)$ returns the route (L7, Fig.~\ref{fig:problem_example}).
It compares the vector costs component by component from the first, so a lower cost at a higher level decides the route.
The search visits the map itself, so the route it returns exists on $G$ and reaches $v_\mathrm{g}$ whenever one does.
\section{Commonsense Planning Benchmark}
\label{sec:benchmark}

A benchmark for commonsense planning needs three things: situations that need commonsense, problems where the right route changes, and a right route on record for each.
Our \emph{commonsense planning benchmark} prepares such situations, problems, and their right routes, with the instructions at different levels of abstraction.

\subsection{Environments}
\label{sec:environments}

Existing indoor environments are scans of real spaces or layouts built for household tasks~\cite{savva2019habitat,ramakrishnan2021hm3d,deitke2022procthor}, so they do not pose situations that need commonsense.
Benchmarks for social navigation exist, but test the robot only against moving people~\cite{biswas2022socnavbench} or the rules written down~\cite{chen2025lisn}.
We therefore build a 2.5D simulator, where we place such situations at the places of our choice.
From one floor plan each, we generate three environments: a supermarket, a restaurant, and a hospital, each with its own considerations for a service robot.
They differ in the shape of the free space, from the parallel aisles of the supermarket to the lattice of tables in the restaurant, and so in the routes around a situation.

\begin{figure}[t]
    \centering
    \includegraphics[width=\linewidth]{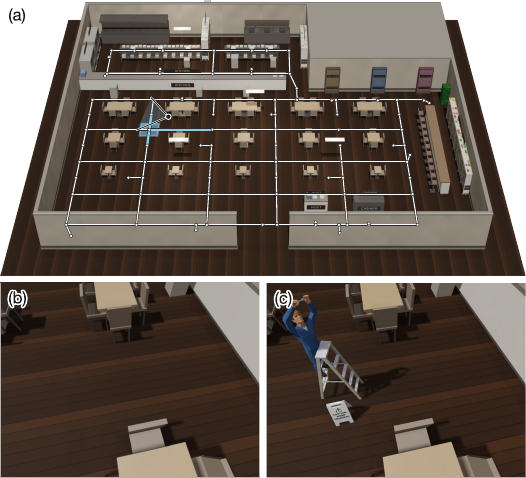}
    \caption{Benchmark construction in the restaurant environment. (a)~A region on the floor, the graph, and the camera that sees it. (b)~The original view from that camera. (c)~The same view with the situation drawn in.}
    \label{fig:benchmark}
\end{figure}

Fig.~\ref{fig:benchmark} shows the restaurant environment with an example situation and its observation image.
The vertices of the map lie at the junctions of the aisles and in front of the landmarks, with the edges along the aisles between them (Fig.~\ref{fig:benchmark}a).
This gives a graph of about 50 to 95 places.
We place the situations as regions on the floor, each a rectangle across an aisle, and an arrangement holds one to three of them.
Each place has a camera that looks down at the floor, so a region appears in the images from the places around it.
Writing each situation as text gives us any case we need, including those that call for complex commonsense.
We use Gemini 3.1 Flash Image to draw them into the images (Fig.~\ref{fig:benchmark}b, c).

\subsection{Problem Types}
\label{sec:problem_types}

The right route changes with the situation, the instruction, the task, and the remaining routes, so each problem type below varies one of these four factors.

\begin{itemize}
    \item \textbf{Unstated care.} The situation changes from one worth a detour to one worth none, so the robot must choose a detour or a crossing from the situation alone.
    \item \textbf{Situational urgency.} The instruction changes from no hurry to a delivery due in minutes.
    \item \textbf{Robot duty.} The duty changes from cleaning the floor with noise to carrying items quietly, so the robot avoids a different region in each problem.
    \item \textbf{Available detour.} The three conditions differ in whether a route around the situation exists, and whether to take it when it does.
\end{itemize}

Every problem puts a situation on the shortest route and makes a detour cost more than a little, so the right route differs from the shortest one.
A problem instance fixes an arrangement of the situations, a pair of endpoints, and an instruction.
Each environment carries nine arrangements and ten pairs of endpoints in each.
Available detour then holds 450 problems across the three environments, and each of the other types holds 300.

\subsection{Levels of Abstraction}
\label{sec:abstraction}

A planner also has to supply what the instruction leaves out, so the instructions come on three levels that differ in how much they state.
The \textbf{coarse} level states the task with its urgency or duty, but nothing about how to move.
The \textbf{base} level adds a single word such as ``safely,'' in three phrasings of the instruction.
The \textbf{detailed} level names the behavior for each situation, along with its object where the behavior has one, as in ``keep off the tactile paving.''
The coarse and base levels thus leave the regions to avoid for the robot to judge, while the detailed level names them and sets a reference.
Across the three environments, the evaluation covers 1350 problems that every planner solves.

\subsection{Ground Truth and Metrics}
\label{sec:metrics}

A metric needs one right route for every problem.
For every arrangement, we therefore rank the regions and define which of them the robot should avoid, and we derive the reference route from that ranking.
The metrics below compare the returned route with this reference.

\begin{itemize}
    \item \textbf{Behavior match.} The fraction of problems in which the route agrees with the reference on all three counts: which regions to avoid, which to cross, and whether to detour.
    \item \textbf{Critical violation.} Among the problems with a critical region, the fraction in which the route crosses it. A critical region is one the robot must not pass under any instruction, such as a worker on a ladder or an infant on the floor.
    \item \textbf{Over-conservatism.} Among the problems where no detour is right, the fraction in which the route takes a detour.
\end{itemize}

Per environment, behavior match covers 450 problems, critical violation 200, and over-conservatism 150.
\section{Experiments}
\label{sec:experiments}

We designed our experiments on the benchmark in Section~\ref{sec:benchmark} to answer the following questions:
\begin{itemize}
    \item[Q1.] Can a planner derive the unstated considerations?
    \item[Q2.] Does a planner return a different route under the same situation when the instruction, the task, or the remaining routes change?
    \item[Q3.] Does turning the judgment into costs on the map return the right route more often than writing or selecting one?
    \item[Q4.] Does ranking the regions against one another tell them apart better than scoring each on its own?
\end{itemize}

\begin{table*}[t!]
  \centering
  \begin{threeparttable}
    \caption{Quantitative benchmark results of the baselines and CoRS by environment}
    \label{tab:main}
    \setlength{\tabcolsep}{5pt}
    \begin{tabular}{l|ccc|ccc|ccc}
      \toprule
      & \multicolumn{3}{c|}{Behavior match $\uparrow$} & \multicolumn{3}{c|}{Critical violation $\downarrow$} & \multicolumn{3}{c}{Over-conservatism $\downarrow$} \\
      \cmidrule(lr){2-4} \cmidrule(lr){5-7} \cmidrule(lr){8-10}
      Planner & Supermarket & Restaurant & Hospital & Supermarket & Restaurant & Hospital & Supermarket & Restaurant & Hospital \\
      \midrule \midrule
      Shortest path           & 35.6 & 41.1 & 34.4 & 85.0 & 65.0 & 65.0 & 0.0 & 0.0 & 0.0 \\
      Uniform caution         & 44.4 & 44.4 & 44.4 & 25.0 & 25.0 & 25.0 & 66.7 & 66.7 & 66.7 \\
      Oracle ranking & 66.7 & 66.7 & 66.7 & 0.0 & 0.0 & 0.0 & 66.7 & 66.7 & 66.7 \\
      \midrule
      Direct planning     & 38.4 & 40.4 & 34.0 & 58.0 & 52.0 & 64.5 & 20.0 & 24.7 & 5.3 \\
      Candidate selection & 61.6 & 61.8 & 57.8 & 31.5 & 43.0 & 36.5 & \textbf{10.0} & \textbf{0.7} & \textbf{0.0} \\
      Absolute scoring    & 64.7 & 70.7 & 64.4 & 24.0 & 0.5 & 9.5 & 38.0 & 45.3 & 72.0 \\
      \rowcolor[gray]{0.9}
      \textbf{Ours: CoRS} & \textbf{76.0} & \textbf{79.6} & \textbf{81.3} & \textbf{6.5} & \textbf{0.0} & \textbf{0.0} & 36.0 & 30.0 & 48.7 \\
      \bottomrule
    \end{tabular}
    \begin{tablenotes}[flushleft]
      \item All values in \%. The planners with an LLM solve each arrangement under all five instructions, while the three without one solve it once. Best among the LLM-based planners in bold.
    \end{tablenotes}
  \end{threeparttable}
\end{table*}

\begin{figure*}[t!]
    \centering
    \includegraphics[width=\linewidth]{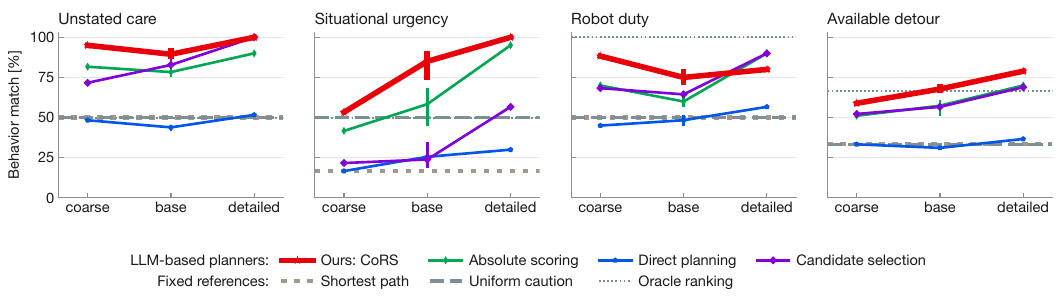}
    \caption{Planner performance across the problem types and the levels of abstraction, in behavior match pooled over the three environments. 
    At base, the marker is the mean across the three wordings, and the bar is their range.}
    \label{fig:problem_types}
\end{figure*}

\subsection{Baselines}
\label{sec:baselines}
We compare CoRS with three planners that differ in how they employ LLMs and VLMs in planning.
\begin{itemize}
    \item \textbf{Direct planning}~\cite{chen2024mapgpt,zhou2024navgpt,zhou2024navgpt2}. 
    The model writes the route from the texts and an adjacency list of the map with neutral node names.
    The route may be absent from the map or miss the goal, in which case the planner falls back to the shortest path.
    \item \textbf{Candidate selection}~\cite{song2025guide,fang2026obstacles}.
    The model selects one of $m$ candidate routes that include the shortest one and differ from one another in shape.
    The candidates come from geometry alone, so the right route may not be among them.
    \item \textbf{Absolute scoring}~\cite{sugino2026vang}.
    The model scores each region on its own and scales the length inside the region by the score.
    No comparison enters the score, so the same region carries the same cost in any situation.
\end{itemize}
We also compare with \textbf{Shortest path}, which searches on the length alone, and \textbf{Uniform caution}, which places every region above the length without reading the texts.
\textbf{Oracle ranking} takes the ranking we define in Section~\ref{sec:metrics} and keeps every region above the length.
It thus shows the gain from the right ranking alone, without the judgment against the length.

\subsection{Implementation Details}
\label{sec:implementation_details}

\paragraph{Model configuration}
We query Claude Haiku 4.5 for every judgment in CoRS and the three baselines that employ foundation models, with no extended thinking and one sample per query.
Claude Sonnet 5 writes the text of each region once from its observation images, and describes only the observation.

\paragraph{CoRS implementation}
CoRS searches with A$^*$~\cite{hart1968formal}, as do the baselines except Direct planning.
The sort is a quicksort with a fixed seed, and it queries the model once for each pair it compares.
The prompt names the regions by neutral labels, so the model reads only the texts.
The search then sums the vector cost in \eqref{eq:ranking_cost} along the edges and orders the open list by it.
The heuristic is the Euclidean distance to the goal on the component $k_{\mathrm{len}}$ and zero elsewhere, which is admissible and consistent.

\paragraph{Baseline implementation}
Every planner reads the same instruction and the same texts of the regions.
Direct planning makes one call to the model per problem.
Candidate selection also makes one call and picks $m = 3$ routes farthest apart in Hausdorff distance among the routes through every place.
Absolute scoring makes one call per region and scores it $s$ from 1 to 5.
The score scales the length inside the region by $6^{z}$ with $z = (s - 3)/2$, which adapts the cost of~\cite{sugino2026vang} to the longer edges of our map.

\subsection{Results}
\label{sec:results}

\paragraph{Comparison with baselines}
\label{sec:quantitative}

We first examine in Table~\ref{tab:main} whether each planner avoids the right regions in each environment, neither too few nor too many.
The three planners without an LLM mark the two extremes.
Shortest path crosses a critical region in most of the problems, while Uniform caution and Oracle ranking take a needless detour in 66.7\%.
CoRS balances the regions against the length, so it violates a critical region in at most 6.5\% and takes far fewer needless detours.
As a result, it reaches the highest behavior match in every environment.
Although Oracle ranking has the regions in the right order, its behavior match remains lower than CoRS, since it keeps every region above the length.

To answer Q1 and Q2, we examine Fig.~\ref{fig:problem_types}, which shows behavior match by problem type and level of abstraction.
In Unstated care, the coarse and base instructions leave out the regions to avoid, so a planner must choose a detour or a crossing from the situation alone.
CoRS reaches the highest behavior match of 95\% with the coarse instruction.
It grounds the instruction on the observation and derives the consideration itself (Q1).

In the other three types, the situation stays the same while the deadline, the duty, or the detour changes the right route.
A planner blind to these changes would return the same route in every condition.
Behavior match is then at most 50\%, or 33\% with the three conditions of Available detour.
CoRS reaches a better behavior match in every type with every instruction.
It ranks the regions and the length under each instruction, so the same arrangement returns a different route in each condition (Q2), as the case studies below show.

To answer Q3 and Q4, we compare CoRS in Table~\ref{tab:main} with the three planners that employ an LLM.
Direct planning has the lowest behavior match, since it writes an invalid route in 67\% of the problems and falls back to the shortest path.
Candidate selection remains low in Situational urgency even with the detailed instruction (Fig.~\ref{fig:problem_types}), since its candidates come from geometry alone.
Absolute scoring comes closest to CoRS, as it also turns the judgment into costs for planning.
Yet its scores come out high and alike, too flat to order the regions.
The planner then spends its detour on a region it could cross, and crosses the one worth a detour.
CoRS thus returns the right route more often than a planner that writes or selects one (Q3).
Its ranking also separates the regions that absolute scores leave alike (Q4).

\begin{figure*}[t]
  \centering
  \subfloat[Hospital, detour available]{\includegraphics[width=0.32\textwidth]{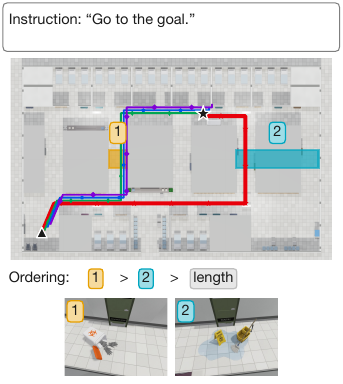}\label{fig:case_a}}\hfill
  \subfloat[Restaurant, noisy robot]{\includegraphics[width=0.32\textwidth]{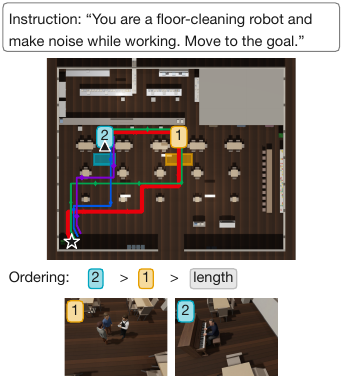}\label{fig:case_b}}\hfill
  \subfloat[Supermarket, urgent]{\includegraphics[width=0.32\textwidth]{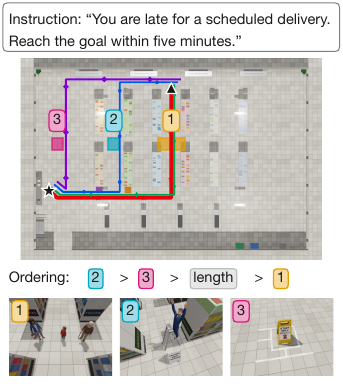}\label{fig:case_c}}\\
  \subfloat[Hospital, no detour]{\includegraphics[width=0.32\textwidth]{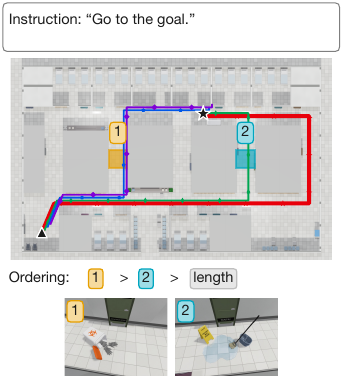}\label{fig:case_d}}\hfill
  \subfloat[Restaurant, quiet robot]{\includegraphics[width=0.32\textwidth]{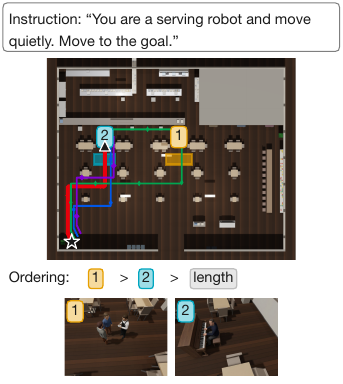}\label{fig:case_e}}\hfill
  \subfloat[Supermarket, no hurry]{\includegraphics[width=0.32\textwidth]{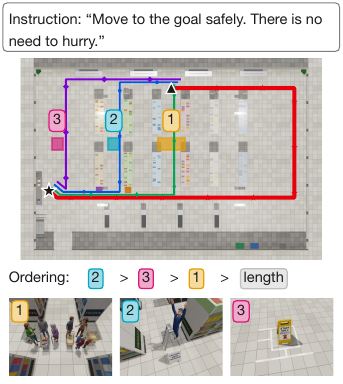}\label{fig:case_f}}\\[5pt]
  \includegraphics[width=\textwidth]{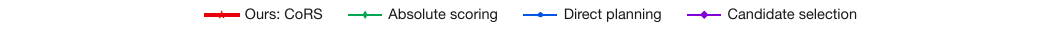}\\[-5pt]
  \caption{
  Case studies. Each panel shows one problem, with the routes of the LLM-based planners over the map ($\blacktriangle$ start, $\bigstar$ goal), the ranking CoRS returned, and the image of each numbered region. The two panels in a column share the start and the goal. Where routes coincide, the baselines are drawn slightly offset; CoRS is drawn on the map's own geometry.}
  \label{fig:case}
\end{figure*}

\paragraph{Case studies} 
\label{sec:qualitative}

Fig.~\ref{fig:case} shows three problem pairs, where the two panels differ in the remaining routes, the robot's duty, or the task urgency.
Each panel shows the routes of CoRS and the three baselines using an LLM, with the observation images below.
In the hospital of Fig.~\ref{fig:case_a} and \ref{fig:case_d}, the instruction says only ``go to the goal.''
CoRS avoids both regions when a route around them exists.
Otherwise, it crosses the wet floor and keeps off the fallen hazardous box, which stands higher in the ranking.
In the restaurant of Fig.~\ref{fig:case_b} and \ref{fig:case_e}, the duty in the instruction changes from cleaning with noise to serving quietly.
The noisy robot then avoids the person playing the piano and passes the two customers talking with a waiter, while the quiet robot does the opposite.
In the supermarket of Fig.~\ref{fig:case_c} and \ref{fig:case_f}, the instruction changes from an urgent delivery to no hurry.
CoRS avoids the worker on a ladder and the staff-only area under both instructions.
The robot instead passes the customers in the aisle to arrive in time, while it takes a longer route that avoids them without the deadline.

\begin{table}[t!]
  \centering
  \begin{threeparttable}
    \caption{Ablation of CoRS over all environments}
    \label{tab:ablation}
    \setlength{\tabcolsep}{4pt}
    \begin{tabular}{l ccc}
      \toprule
      & Behavior & Pair & Over- \\
      & match $\uparrow$ & agreement $\uparrow$ & conservatism $\downarrow$ \\
      \midrule
      \rowcolor[gray]{0.9}
      \textbf{Ours: CoRS}       & 79.0 & 75.5 & 38.2 \\
      \midrule
      Source descriptions & 95.3 & 91.6 & 5.6 \\
      \midrule
      Low thinking        & 77.2 & 78.0 & 40.3 \\
      High thinking       & 76.3 & 75.1 & 40.7 \\
      \midrule
      Sonnet 5            & 80.8 & 79.2 & 17.6 \\
      Opus 5              & 76.3 & 89.6 & 42.5 \\
      \bottomrule
    \end{tabular}
    \begin{tablenotes}[flushleft]
        \item All values in \%. CoRS uses Haiku 4.5 for the judgments, as in Table~\ref{tab:main}. The rows below replace the descriptions, the thinking budget, and the judgment model in turn.
    \end{tablenotes}
  \end{threeparttable}
\end{table}

\begin{figure}[t]
  \centering
  \subfloat[A crowded aisle]{\includegraphics[width=0.32\linewidth]{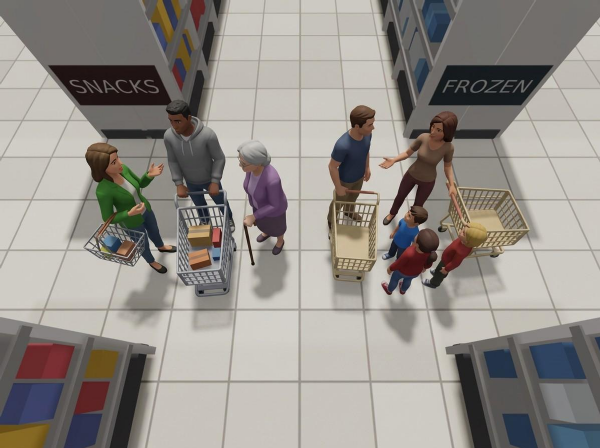}\label{fig:omit_blocked}}\hfill%
  \subfloat[A sparse aisle]{\includegraphics[width=0.32\linewidth]{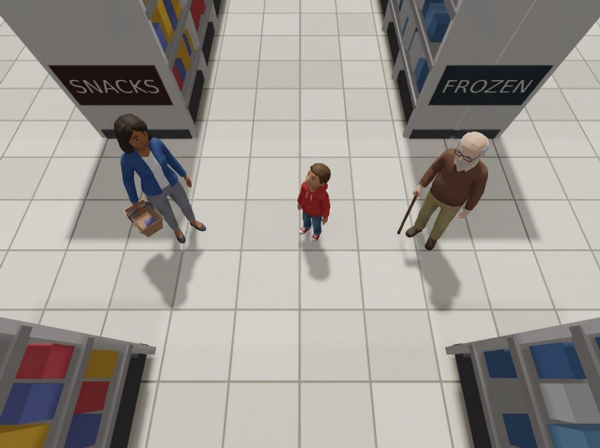}\label{fig:omit_sparse}}\hfill%
  \subfloat[A dry floor]{\includegraphics[width=0.32\linewidth]{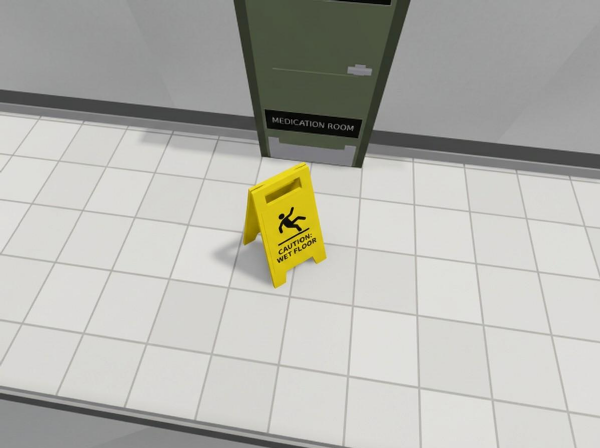}\label{fig:omit_dry}}
  \caption{Example observations where the description misses a fact about the space around the objects on the floor.}
  \label{fig:omissions}
\end{figure}

\paragraph{Ablations}
\label{sec:ablation}

We run ablations of CoRS in Table~\ref{tab:ablation}, which replace one part of the judgment.
Source descriptions give the LLM the texts we wrote to render the images without a VLM.
The rest vary the thinking budget and the model, which both shape the quality of each judgment.
For the ablations, we also read the pairwise judgments behind the ranking in Section~\ref{sec:metrics}.
Pair agreement is the fraction of the comparisons that agree with them.

We first look at source descriptions, which give the largest gain.
While VLM-generated descriptions remain neutral on the judgment as intended, they often miss the facts behind one. 
For example, they capture the people and carts blocking an aisle (Fig.~\ref{fig:omit_blocked}), but miss the navigable gaps between shoppers (Fig.~\ref{fig:omit_sparse}).
They also read a cleaning sign but not the dry tiles under it (Fig.~\ref{fig:omit_dry}), so CoRS reads the sign as a wet floor and detours, while the source text crosses.

A stronger model and a larger thinking budget both supply the hazards the description leaves out. 
On the same dry floor, Opus 5 assumes the description misses a film of water too thin to see and takes a detour twice as long. 
It agrees with our reference best, yet it detours around 91\% of the regions, against 82\% for CoRS.
A larger thinking budget turns the judgment from the people on the floor to the robot's own safety. 
In one problem, the robot avoids a puddle rather than a person crouching beside a fallen cane.

\subsection{Limitations and Future Directions}
\label{sec:limitation}

Although CoRS takes far fewer needless detours than Absolute scoring, it still has room to rank a region against the detour around it.
As the ablations suggest, a stronger model would plan better if it did not assume beyond the observation.
Another limit is that CoRS judges the regions once before the robot moves, so a situation that changes on the way falls outside the ranking.
Running the judgment online would need local observations, and the ranking could then feed the lower-level controller.

\section{Conclusion}
\label{sec:conclusion}

We presented CoRS, a path planner that turns an abstract instruction into a route that follows commonsense.
It infers the latent considerations from each region, ranks the regions against one another, and searches for a valid route on the resulting costs.
The same region is thus avoided in one situation and crossed in another, as the instruction, the task, and the remaining routes change.
On a benchmark of simulated everyday environments, CoRS followed commonsense more often than every baseline.
Future work will integrate CoRS into a navigation system with local observation and control, and deploy it in the real world.


\bibliographystyle{IEEEtran}
\bibliography{venues_abbrev, reference}

\end{document}